\documentclass[conference]{IEEEtran}
\IEEEoverridecommandlockouts

\usepackage{cite}
\usepackage{amsmath,amssymb,amsfonts}
\usepackage{algorithmic}
\usepackage{graphicx}
\usepackage{textcomp}
\usepackage{xcolor}
\usepackage{hyperref}
\usepackage{multirow}
\usepackage{url}
\usepackage{booktabs}
\usepackage{tabularx}
\usepackage{CJKutf8}
\def\BibTeX{{\rm B\kern-.05em{\sc i\kern-.025em b}\kern-.08em
    T\kern-.1667em\lower.7ex\hbox{E}\kern-.125emX}}
\begin{document}

\title{Video-guided Machine Translation \\ with Global Video Context}


\author{\IEEEauthorblockN{1\textsuperscript{st} Jian Chen}
\IEEEauthorblockA{
\textit{Shenzhen University}\\
2310413006@email.szu.edu.cn}
\and
\IEEEauthorblockN{2\textsuperscript{nd} JinZe Lv}
\IEEEauthorblockA{
\textit{Shenzhen University}\\
lvjinze777@163.com}
\and
\IEEEauthorblockN{3\textsuperscript{rd} Zi Long}
\IEEEauthorblockA{
\textit{Shenzhen Technology University}\\
longzi@sztu.edu.cn}
\and
\IEEEauthorblockN{4\textsuperscript{th} XiangHua Fu}
\IEEEauthorblockA{
\textit{Shenzhen Technology University}\\
fuxianghua@sztu.edu.cn}
}

\maketitle

\begin{abstract}
Video-guided Multimodal Translation (VMT) has advanced significantly in recent years. However, most existing methods rely on locally aligned video segments paired one-to-one with subtitles, limiting their ability to capture global narrative context across multiple segments in long videos. To overcome this limitation, we propose a globally video-guided multimodal translation framework that leverages a pretrained semantic encoder and vector database-based subtitle retrieval to construct a context set of video segments closely related to the target subtitle semantics. An attention mechanism is employed to focus on highly relevant visual content, while preserving the remaining video features to retain broader contextual information. Furthermore, we design a region-aware cross-modal attention mechanism to enhance semantic alignment during translation. Experiments on a large-scale documentary translation dataset demonstrate that our method significantly outperforms baseline models, highlighting its effectiveness in long-video scenarios.
\end{abstract}


\section{Introduction}
Neural machine translation (NMT) has achieved remarkable progress on text-only tasks~\cite{bahdanau2014neural,wu2016google}. However, source texts often suffer from sparsity or ambiguity, hindering accurate translation. Multimodal machine translation (MMT) addresses this by incorporating auxiliary modalities like images or videos to enhance semantic understanding~\cite{specia2016shared,sulubacak2020multimodal}.

Early MMT primarily leveraged static images for semantic alignment~\cite{elliott2016multi30k,tang2022multimodal,long2024exploring}, integrating visual features during encoding~\cite{calixto2017incorporating,huang2016attention} or decoder initialization~\cite{elliott2015multilingual,madhyastha2017sheffield}. Recently, video-guided multimodal translation (VMT)~\cite{wang2019vatex,li2022visa} has gained attention, as videos provide richer temporal and spatial cues essential for contextual coherence.

Despite these advances, existing VMT methods often rely on local alignment~\cite{gu2021video,kang2023bigvideo}, neglecting the global narrative context crucial for long-form videos. To address this, recent works like TopicVD~\cite{lv2025topicvd} introduced global context from documentaries. Building on this, we propose a global context-aware framework that integrates distributed visual information to improve translation accuracy beyond local alignment.

Our main contributions are:
\begin{itemize}
    \item We retrieve subtitle-relevant video segments using semantic retrieval augmented by a context fusion strategy preserving temporal adjacency.
    \item We introduce a Cross-modal Attention mechanism to adaptively select critical segments, fusing unselected ones to mitigate semantic loss.
    \item We propose a Region-aware Cross-modal Attention module to selectively aggregate region-level visual semantics via hierarchical token-region interactions.
\end{itemize}

\begin{figure*}[h]
    \centering
    \includegraphics[scale=0.34]{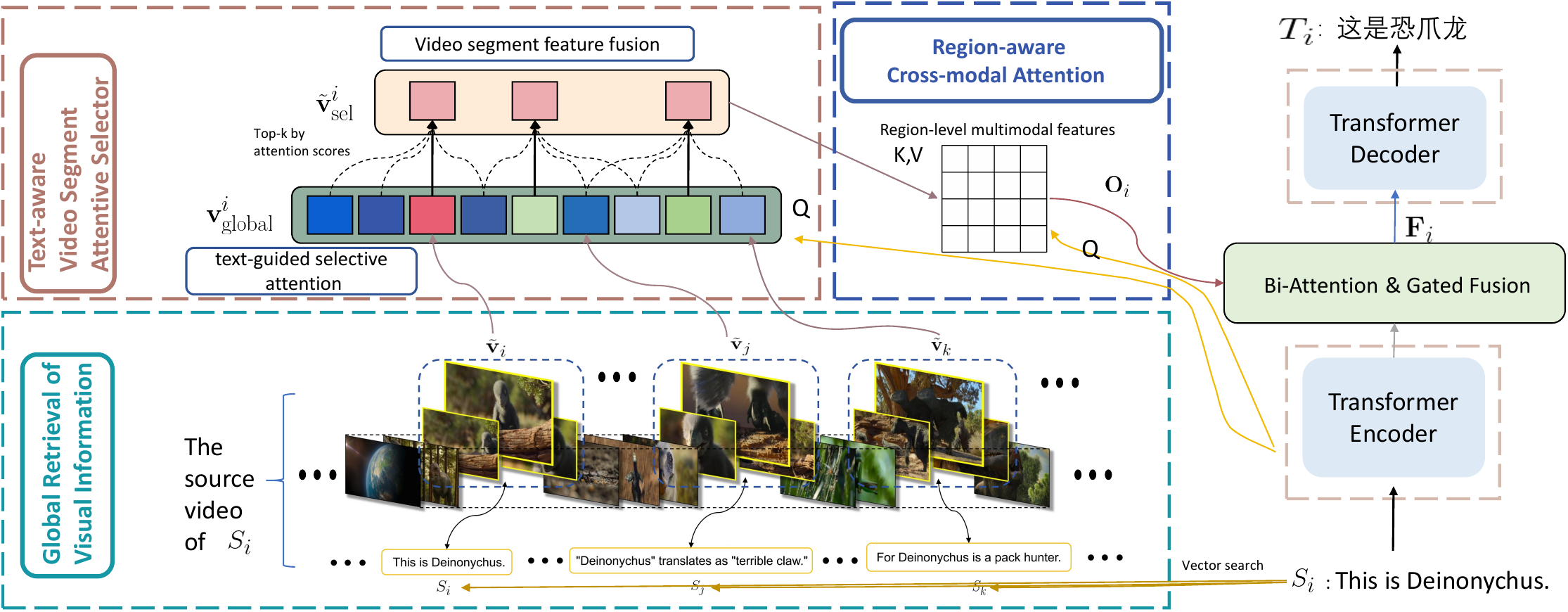} 
    \caption{The Overview of the Framework of the Proposed Method}
    \label{fig:framework}
\end{figure*}


\section{Related Work}

Attention mechanisms are central to MMT for integrating visual information. Caglayan et al.~\cite{caglayan2016multimodal} and Calixto et al.~\cite{calixto2017doubly} proposed multimodal attention modules and doubly-attentive decoders, respectively, to dynamically regulate visual feature interaction. Tang et al.~\cite{tang2022multimodal} further extended MMT to text-only datasets via retrieval-based frameworks.

VMT extends MMT by capturing temporal dynamics. Li et al.~\cite{li2023video} and Gu et al.~\cite{gu2021video} introduced attention mechanisms to resolve subtitle ambiguities and model dynamic spatial-temporal features. Kang et al.~\cite{kang2023bigvideo} improved alignment via cross-modal contrastive learning on large-scale data. However, existing methods mostly rely on local alignment, neglecting global narrative cues essential for long-form videos.

Early datasets like How2~\cite{sanabria2018how2} and VATEX~\cite{wang2019vatex} provide large-scale pairs but often suffer from text-dependency. Smaller datasets like VISA~\cite{li2022visa} focus on disambiguation but lack robustness. While recent corpora such as BigVideo~\cite{kang2023bigvideo} and EVA~\cite{li2023video} exceed one million samples, they primarily contain short, simple captions. To address this, TopicVD~\cite{lv2025topicvd} introduces topic-aware, document-derived samples to facilitate global context modeling.

\section{The Proposed Method}
Fig.~\ref{fig:framework} illustrates the overall architecture of our proposed framework. As depicted, the model comprises five key components: Global Visual Information Retrieval, a Text-Aware Video Segment Selector, Region-Aware Cross-Modal Attention, Bi-Directional Attention with Gated Fusion and Transformer-based Text Generation~\cite{vaswani2017attention}.

\subsection{Global Retrieval of Visual Information}
\label{sec:Construction}

Given a documentary video \(D = \{(S_i, T_i), V_i\}_{i=1}^N\), where \(N\) denotes the number of subtitles in \(D\), each \(S_i\) represents the \(i\)-th subtitle, \(T_i\) is its human translation, and \(V_i\) denotes the corresponding video segment. In our setting, \(S_i\) serves as the source input text, while \(T_i\) is used as the target output text for translation.

To retrieve global visual information for a given sample \((S_i, V_i)\), we first embed the query subtitle \(S_i\) into a high-dimensional semantic space using a pretrained text encoder. Based on this embedding, we retrieve the top-\(P\) subtitles from the video \(D\) whose representations are most semantically similar to \(S_i\):
\begin{equation}
\mathcal{I}_{\text{global}}^i = \arg\max_x \mathrm{Top}(\mathrm{sim}(S_x, S_i), P)
\end{equation}
where \(1 \leq x \leq N\), \(\mathrm{sim}(S_x, S_i)\) denotes the semantic similarity between the subtitle \(S_x\) and the query subtitle \(S_i\), and \(\mathrm{Top}(l, n)\) denotes an operation that selects the \(n\) largest values from the list \(l\). Accordingly, \(\mathcal{I}_{\text{global}}^i = \{p_1, p_2, \dots, p_P\}\) represents the indices of the \(P\) subtitles that are most semantically similar to \(S_i\) in the entire video.

We denote the video segments corresponding to the retrieved relevant subtitle indices as \(\mathbf{R}_{\text{global}}^i = \{V_j \mid j \in \mathcal{I}_{\text{global}}^i \}\). To further enrich the contextual information, we exploit the temporal locality commonly observed in documentary-style videos: neighboring segments often describe related events or entities. For each selected segment \(V_j\), we aggregate the feature vectors of its \(w\) preceding and \(w\) succeeding adjacent segments, scaled by a fixed factor \(\gamma\) to reduce semantic noise. The resulting enhanced segment representation is computed as:
\begin{equation}
\tilde{\mathbf{v}}_j = \mathbf{v}_j + \gamma \cdot \sum_{k \in \mathcal{N}_w(j)} \mathbf{v}_k, \quad j \in \mathcal{I}_{\text{global}}^i,
\end{equation}
where \(\mathcal{N}_w(j)\) denotes the indices of segments adjacent to segment \(j\) within a window of size \(w\) and \(\mathbf{v}_j\) is the feature vector of \(V_j\).
The final globally contextualized video set is then constructed by collecting the enhanced features of all selected segments:
\begin{equation}
\mathbf{V}_{\text{global}}^i = \left\{ \tilde{\mathbf{v}}_j \mid j \in \mathcal{I}_{\text{global}}^i \right\}.
\end{equation}

This process effectively identifies subtitle-relevant video segments within potentially redundant content, leveraging a semantic retrieval strategy augmented by context fusion from neighboring segments. By incorporating adjacent segment information, the method preserves temporal and contextual cues, thereby maintaining semantic consistency along the video’s temporal dimension.

\subsection{Text-aware Video Segment Attentive Selector}
\label{sec:VCSF}

To fully exploit global video context and enhance semantic richness, we propose a Text-aware Video Segment Attentive Selector. This module accepts the contextualized video set \(\mathbf{V}_{\text{global}}^i\) as input, adaptively selects semantically salient segments, and integrates information from unselected segments to mitigate information loss.

Given a source subtitle \(S_i\), encoded as \(\mathbf{T}_i \in \mathbb{R}^{L \times E_t}\), we quantify the cross-modal relevance using attention-based alignment scores:
\begin{equation}
\boldsymbol{\alpha} = \mathrm{Softmax}([s_1, \dots, s_P]), \quad \text{where } s_j = \mathrm{Attn}(\mathbf{T}_i, \tilde{\mathbf{v}}_j)
\end{equation}
Here, \(\mathrm{Attn}(\cdot)\) denotes the attention function with \(\mathbf{T}_i\) as the query and \(\tilde{\mathbf{v}}_j \in \mathbf{V}_{\text{global}}^i\) as the key. Subsequently, we select the top-\(K\) relevant video segments to constitute the primary visual context \(\mathbf{V}_{\mathrm{sel}}^i\):
\begin{equation}
\mathcal{I}_{\mathrm{sel}}^i = \arg\max_x \mathrm{Top}(\alpha_x, K)
\end{equation}
\begin{equation}
\mathbf{V}_{\mathrm{sel}}^i = \left\{ \tilde{\mathbf{v}}_j \mid j \in \mathcal{I}_{\mathrm{sel}}^i \right\}
\end{equation}
where \(\mathcal{I}_{\mathrm{sel}}^i\) denotes the indices of the \(K\) segments most relevant to \(S_i\).


To preserve a broader scope of context, we exploit temporal locality by aggregating 
information from the unselected set $\mathbf{V}_u^i$. Specifically, for each selected 
segment $\tilde{\mathbf{v}}_j \in \mathbf{V}_{\mathrm{sel}}^i$, we apply the following 
update rule:
\begin{equation}
\tilde{\mathbf{v}}'_j = \tilde{\mathbf{v}}_j + \frac{\lambda}{2} \sum_{K_{j-1} \leq k 
\leq K_{j+1}, \tilde{\mathbf{v}}_k \in \mathbf{V}^i_{\mathrm{u}}} \tilde{\mathbf{v}}_k
\end{equation}
where $\lambda$ is a fixed weighting coefficient, and $K_j$ denotes the index of 
$\tilde{\mathbf{v}}_j$ in $\mathbf{V}_{\mathrm{global}}^i$, so the summation aggregates 
unselected segments lying between the $(j-1)$-th and $(j+1)$-th selected segments in 
the globally retrieved set. This enriches each selected segment with neighboring context 
while maintaining temporal coherence. The final updated set is denoted as:
\begin{equation}
\tilde{\mathbf{V}}_{\mathrm{sel}}^i = \{ \tilde{\mathbf{v}}'_j \mid j \in 
\mathcal{I}_{\mathrm{sel}}^i \}
\end{equation}

Distinct from the static construction in Section~\ref{sec:Construction}, this mechanism is optimized during training, facilitating the dynamic identification of semantically relevant segments. Furthermore, by incorporating a context-aware fusion strategy, the approach mitigates local semantic loss, preserving a coherent visual context for downstream multimodal translation.

\subsection{Region-Aware Cross-modal Attention}

To facilitate fine-grained alignment between textual and visual region-level features, we propose a Region-aware Cross-modal Attention module. This module enables deep cross-modal fusion via hierarchical semantic interactions.


Let $\tilde{\mathbf{V}}_{\mathrm{sel}}^i \in \mathbb{R}^{K \times R \times E_v}$ denote the visual features, where $K$, $R$, and $E_v$ represent the number of video segments, spatial regions, and feature dimensions, respectively. Here, $R$ denotes the number of spatial grid cells in the I3D feature maps, each corresponding to a local region of the video frame. To unify feature spaces, we first project the textual features $\mathbf{T}_i$ into the visual space via a learnable linear transformation:
\begin{equation}
\mathbf{T}_i' = \mathcal{W}_t \mathbf{T}_i + \mathbf{b}_t, \quad \mathbf{T}_i' \in \mathbb{R}^{L \times E_v}
\end{equation}
Subsequently, $\mathbf{T}_i'$ is expanded to $\mathbb{R}^{R \times L \times E_v}$ by repeating along the region dimension, and $\tilde{\mathbf{V}}_{\mathrm{sel}}^i$ is reshaped to $\mathbb{R}^{R \times K \times E_v}$, enabling each spatial region to independently attend to the full text sequence.

We employ multi-head attention where queries are derived from textual features, and keys/values from video features. The attended representation \(\mathbf{Z}_i\) is computed via scaled dot-product attention:
\begin{equation}
\mathbf{Z}_i = \mathrm{Softmax}\left( \frac{(\mathcal{W}_q \mathbf{T}_i') (\mathcal{W}_k \tilde{\mathbf{V}}_{\mathrm{sel}}^i)^\top}{\sqrt{d_k}} \right) (\mathcal{W}_v \tilde{\mathbf{V}}_{\mathrm{sel}}^i)
\end{equation}

Finally, we apply average pooling over the region dimension \(R\), followed by an output projection to yield the fused representation \(\mathbf{O}_i \in \mathbb{R}^{L \times E_o}\):
\begin{equation}
\mathbf{O}_i = \mathcal{W}_o \left( \frac{1}{R} \sum_{r=1}^R \mathbf{Z}_i \right) + \mathbf{b}_o
\end{equation}
where \(\mathcal{W}_o\) and \(\mathbf{b}_o\) are parameters of the output projection layer. This mechanism explicitly integrates regional visual semantics into textual representations, effectively bridging the modality gap.

\subsection{Bi-directional Cross-modal Attention with Gated Fusion}

To facilitate deep semantic interaction, we introduce a Bi-directional Cross-modal Attention mechanism~\cite{lu2019vilbert,su2021multi} that simultaneously captures Text-to-Video (T2V) and Video-to-Text (V2T) dependencies.

In the T2V direction, textual features \(\mathbf{T}_i\) query the video features \(\mathbf{O}_i\) to obtain the context-aware representation \(\mathbf{H}_{\textbf{t2v}}\):
\begin{equation}
\mathbf{H}_{\textbf{t2v}} = \mathrm{Softmax} \left( \frac{\mathbf{T}_i \cdot \mathbf{O}_i^\top}{\sqrt{d_h}} \right) \cdot \mathbf{O}_i
\end{equation}

Conversely, in the V2T direction, video features \(\mathbf{O}_i\) serve as queries to attend to textual features \(\mathbf{T}_i\):
\begin{equation}
\mathbf{H}_{\textbf{v2t}} = \mathrm{Softmax} \left( \frac{\mathbf{O}_i \cdot \mathbf{T}_i^\top}{\sqrt{d_h}} \right) \cdot \mathbf{T}_i
\end{equation}

To integrate these representations, we employ a gated fusion mechanism~\cite{yin2020novel,lin2020dynamic} that adaptively balances visual and linguistic information. The final fused representation \(\mathbf{F}_i\) is formulated as:
\begin{equation}
\mathbf{F}_i = (1 - \mathbf{g}) \odot \mathbf{H}_{\textbf{v2t}} + \mathbf{g} \odot \mathbf{H}_{\textbf{t2v}} + (1 - \mathbf{g}) \odot \mathbf{T}_i
\end{equation}
where \(\odot\) denotes element-wise multiplication. The gating matrix \(\mathbf{g}\) is a learnable parameter that dynamically modulates the relative contribution of each cross-modal pathway during fusion.

\subsection{Transformer-based Text Generation}

The fused representation \(\mathbf{F}_i\) encapsulates bidirectional cross-modal contextual information and serves as the multimodal input to the Transformer decoder, facilitating joint attention to both linguistic and visual semantics throughout the generation process. At each decoding step \(t\), the decoder interacts with \(\mathbf{F}_i\) via multi-head cross-attention, formulated as:
\begin{equation}
\mathbf{C}_t = \mathrm{Softmax}\left(\frac{\mathbf{Q}_{\text{dec}}(t) \mathbf{F}_i^\top}{\sqrt{d_k}}\right) \mathbf{F}_i
\end{equation}
where \(\mathbf{Q}_{\text{dec}}(t)\) denotes the decoder's evolving query vector, and \(\mathbf{F}_i\) functions as unified key-value pairs. The resulting contextual vector \(\mathbf{C}_t\) is subsequently processed by the decoder’s feed-forward layers to produce the hidden state \(\mathbf{h}_t\).

The generation mechanism proceeds autoregressively through sequential token prediction. Specifically, at decoding step \(t\), the hidden state \(\mathbf{h}_t\) is projected linearly onto the target vocabulary space via \(\mathbf{o}_t = \mathbf{W}_o \mathbf{h}_t + \mathbf{b}_o\), where \(\mathbf{W}_o \in \mathbb{R}^{V \times d_h}\), yielding logits that are normalized by the softmax function to form the conditional probability distribution:
\begin{equation}
P(y_t \mid y_{1:t-1}, \mathbf{F}_i) = \mathrm{Softmax}(\mathbf{o}_t)
\end{equation}
The predicted token \(\hat{y}_t\) is obtained by maximizing this conditional probability, formally expressed as:
\begin{equation}
\hat{y}_t = \arg\max_{y \in \mathrm{vocab}} P(y_t = y)
\end{equation}

This predicted token is then fed forward to generate the subsequent query vector \(\mathbf{Q}_{\text{dec}}(t+1)\) for the next decoding iteration. This process continues iteratively until an end-of-sequence (EOS) token is produced, resulting in the complete translated sequence:
\begin{equation}
\hat{\mathbf{y}} = \{\hat{y}_1, \ldots, \hat{y}_T\}, \quad \hat{y}_T = \langle \mathrm{EOS} \rangle
\end{equation}

\section{Experiments}

\subsection{Dataset}
\label{sec:dataset}
We conduct experiments on the TopicVD dataset~\cite{lv2025topicvd}, comprising 256 documentary videos and 122,930 English--Chinese subtitle pairs. The dataset is split into 102,502 training pairs (219 videos), 10,429 validation pairs (18 videos), and 9,999 testing pairs (19 videos).

To assess generalizability, we also evaluate our method on the BigVideo dataset~\cite{kang2023bigvideo}. To ensure a fair comparison with TopicVD regarding data scale, we utilize a subset containing the first 104,790 training pairs, while keeping the original validation and test splits unchanged.\footnote{We also experimented on other VMT datasets~\cite{li2022visa,teramen2024english,wang2019vatex}. However, due to their lack of rich contextual video information, our method yields performance comparable to existing approaches.}
\begin{table}[!t]
\caption{Performance Comparison on the TopicVD Dataset}
\label{tab:bleu_scores}
\centering
\begin{tabular}{lcc} 
\toprule
\textbf{Method} & \textbf{BLEU} & \textbf{METEOR} \\
\midrule
Text-only NMT & 19.14 & 33.95 \\
Image-MMT & 28.93 & 38.59 \\
Segment-Level Video-MMT & 29.23 & 38.65 \\
BigVideo (Kang et al., 2023) & 25.92 & 36.54 \\
\textbf{Proposed Method} & \textbf{30.47} & \textbf{38.96} \\
\bottomrule
\end{tabular}
\end{table}
\subsection{Baseline Models}
We compare our proposed approach against the following baselines:

\begin{itemize}
  \item \textbf{Text-only NMT:} A standard Transformer-based NMT model~\cite{vaswani2017attention}, implemented via the open-source \texttt{transformer-translator-pytorch} project\footnote{\url{https://github.com/devjwsong/transformer-translator-pytorch}}.
  
  \item \textbf{Image-MMT:} An efficient visually-enhanced baseline where visual features are extracted from sampled frames. We employ FAISS~\cite{johnson2019billion} to retrieve the single most relevant frame for each subtitle as auxiliary input.
  
  \item \textbf{Segment-Level Video-MMT:} Utilizes only the video segment temporally aligned with the target subtitle as visual input, ignoring global context.
  
  \item \textbf{BigVideo (Kang et al.):} We reproduce the method proposed in~\cite{kang2023bigvideo}, which leverages cross-modal contrastive learning to align video and text representations. This serves as a strong baseline for video-guided translation.
\end{itemize}

All models are trained on TopicVD and evaluated using 4-gram BLEU~\cite{papineni2002bleu} and METEOR~\cite{banerjee2005meteor}.

\subsection{Experimental Settings}

\textbf{Model Architecture.} We implement our model using Fairseq~\cite{ott2019fairseq}, following the configuration in~\cite{li2022vision}. The architecture consists of 4 encoder and decoder layers, with a hidden dimension of 128, feed-forward size of 256, 8 attention heads, and a dropout rate of 0.35. Video features are extracted via a pretrained I3D model~\cite{carreira2017quo} and linearly projected to align with text embeddings. The total computational complexity is approximately 1.22 GFLOPs per sample.

\textbf{Hyperparameters.} For global retrieval, we utilize CLIP~\cite{radford2021learning} and FAISS~\cite{johnson2019billion} to select the top \(P=10\) subtitles. Context enrichment employs a window of \(w=2\) with a scaling factor \(\gamma=0.1\). In the selector module, we set \(K=5\) and the fusion weight \(\lambda=0.1\).\footnote{Both \(\gamma\) and \(\lambda\) are set to 0.1 to balance context enhancement against noise. Future work may explore adaptive weighting strategies.}

\textbf{Optimization.} We train using RAdam~\cite{liu2019variance} with a learning rate of 0.001, an inverse square root scheduler, and 4000 warm-up steps. The objective is cross-entropy with label smoothing (0.1) and a batch size of 4096 tokens. Training proceeds for up to 20,000 steps with early stopping after 10 epochs of no improvement. Mixed-precision (FP16) training is adopted for efficiency.

\subsection{Experimental Results}

To ensure robustness, we report average BLEU~\cite{papineni2002bleu} and METEOR~\cite{banerjee2005meteor} scores over three independent runs with distinct random seeds.

\textbf{Results on TopicVD.} As shown in Table~\ref{tab:bleu_scores}, our proposed method achieves the best performance on TopicVD (30.47 BLEU / 38.96 METEOR). It significantly outperforms the Text-only baseline (+11.33 BLEU) and the Image-MMT model (+1.54 BLEU). Crucially, within the video-guided domain, our approach surpasses the Segment-Level Video-MMT and BigVideo~\cite{kang2023bigvideo} baselines by 1.24 and 4.55 BLEU points, respectively, demonstrating the efficacy of global context modeling.

\begin{table}[!t]
\caption{Performance Comparison (BLEU and METEOR) on the BigVideo Dataset}
\label{tab:bigvideo-results}
\centering
\newcolumntype{Y}{>{\centering\arraybackslash}X} 

\begin{tabularx}{\linewidth}{l Y Y} 
\toprule
\textbf{Method} & \textbf{BLEU} & \textbf{METEOR} \\
\midrule
BigVideo (Kang et al., 2023) & 44.44 & --- \\
BigVideo (our implementation) & \textbf{47.05} & \textbf{49.63} \\
Proposed Method & \underline{46.78} & \underline{49.45} \\
\bottomrule
\end{tabularx}
\end{table}

\begin{figure*}[!t]
    \centering
    \includegraphics[scale=0.35]{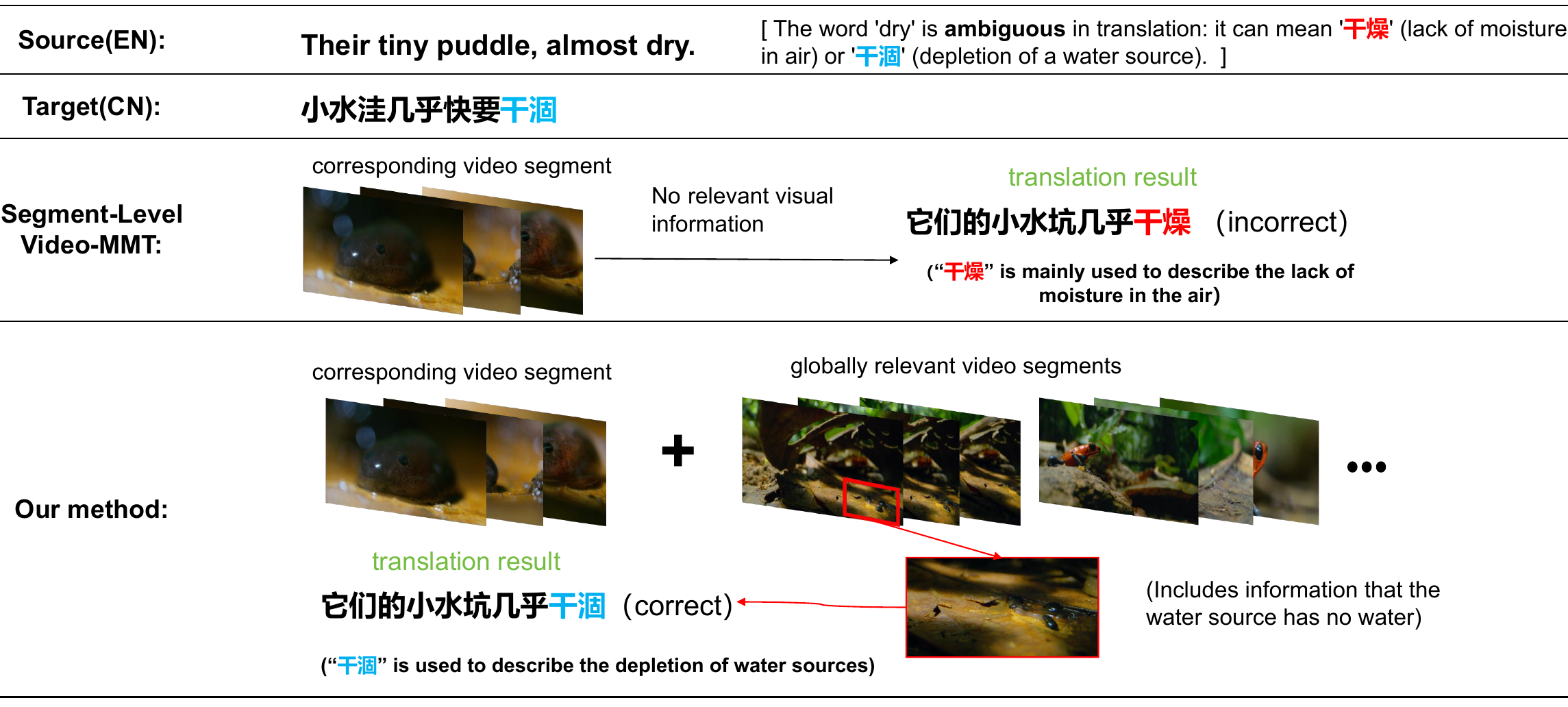}
    \caption{Example of correct translation by the proposed method. The source word ``dry'' is ambiguous in Chinese translation, requiring visual context to disambiguate between \begin{CJK*}{UTF8}{gbsn}``干燥'' \end{CJK*}(lack of moisture in air) and \begin{CJK*}{UTF8}{gbsn}``干涸'' \end{CJK*}(depletion of a water source).}
    \label{fig:example}
\end{figure*}

\textbf{Results on BigVideo.} Table~\ref{tab:bigvideo-results} presents results on the BigVideo subset. Our re-implementation of the BigVideo model achieves 47.05 BLEU. Under the same setting, our method attains a comparable score of 46.78 BLEU.\footnote{The inconsistency between our re-implementation and the original result is attributable to the use of a subset of the BigVideo dataset rather than hyperparameter tuning. For our proposed method, only the retrieval hyperparameters 
$P$, $w$, and $K$ were adjusted to fit this subset.} This performance parity is expected, as BigVideo consists primarily of short segments with limited temporal context, rendering our long-range modeling less advantageous than in documentary scenarios.

\textbf{Qualitative Analysis.} Fig.~\ref{fig:example} illustrates a representative case. The Segment-Level baseline fails to correctly translate ``dry'' due to misleading local visual cues. In contrast, our method successfully resolves this ambiguity by retrieving globally contextualized video segments, providing sufficient semantic evidence to generate an accurate translation.

\begin{table}[!t]
\caption{Ablation Studies on Hyperparameters $P$, $w$, and $K$ on the TopicVD Dataset}
\label{tab:ablation_all}
\centering
\begin{tabular}{lccc} 
\toprule
\textbf{Parameter} & \textbf{Value} & \textbf{BLEU} & \textbf{METEOR} \\
\midrule
\multirow{3}{*}{$P$ ($w=2, K=5$)} 
 & 5 & 29.71 & 38.04 \\
 & 10 & \textbf{30.47} & \textbf{38.96} \\
 & 20 & 30.23 & 38.68 \\
\midrule
\multirow{5}{*}{$w$ ($P=10, K=5$)} 
 & 1 & 29.82 & 38.18 \\
 & 2 & \textbf{30.47} & \textbf{38.96} \\
 & 3 & 30.17 & 38.21 \\
 & 4 & 30.39 & 38.43 \\
 & 5 & 30.06 & 38.03 \\
\midrule
\multirow{6}{*}{$K$ ($P=10, w=2$)} 
 & 3 & 30.14 & 38.21 \\
 & 4 & 30.15 & 38.24 \\
 & 5 & \textbf{30.47} & \textbf{38.96} \\
 & 6 & 29.85 & 38.34 \\
 & 7 & 29.72 & 38.31 \\
 & 10 & 29.51 & 38.22 \\
\bottomrule
\end{tabular}
\end{table}

\section{Analysis and Discussion}

This section analyzes the impact of three critical hyperparameters introduced in Sections~\ref{sec:Construction} and~\ref{sec:VCSF}: retrieval size \(P\), context window \(w\), and selection count \(K\). We further conduct ablation studies to validate the contributions of core modules.

\subsection{Hyperparameter Sensitivity}

Table~\ref{tab:ablation_all} summarizes the performance variations under different hyperparameter settings.

\textbf{Effect of \(P\).} With \(w=2\) and \(K=5\), performance peaks at \(P=10\) (30.47 BLEU). Both smaller (\(P=5\)) and larger (\(P=20\)) values degrade performance, indicating that \(P=10\) strikes an optimal balance between sufficient contextual diversity and noise suppression.

\textbf{Effect of \(w\).} Fixing \(P=10\) and \(K=5\), we find \(w=2\) yields the best results. Narrower windows (\(w=1\)) limit contextual cues, whereas excessively wide windows (\(w=5\)) introduce redundancy and noise, hampering translation accuracy.

\textbf{Effect of \(K\).} With \(P=10\) and \(w=2\), the model achieves the highest scores at \(K=5\). Lower values (\(K<5\)) risk insufficient visual grounding, while higher values (\(K>5\)) tend to incorporate irrelevant or conflicting visual signals.
\begin{table}[!t]
\caption{Ablation Results on the Two Core Modules: GR and TVSS}
\label{tab:ablation}
\centering
\begin{tabularx}{\linewidth}{X c c} 
\toprule
\textbf{Setting} & \textbf{BLEU} & \textbf{METEOR} \\
\midrule
Full Model & \textbf{30.47} & \textbf{38.96} \\
w/o GR & 29.59 & 38.14 \\
w/o TVSS & 29.71 & 38.04 \\
w/o Both Modules & 29.48 & 38.06 \\
\bottomrule
\end{tabularx}
\end{table}
\subsection{Ablation Study}

To assess the efficacy of our proposed components, we conduct an ablation study by systematically removing the Global Retrieval (GR) and Text-aware Video Segment Attentive Selector (TVSS) modules. Results are detailed in Table~\ref{tab:ablation}.

\textbf{Impact of Global Retrieval.} Removing the GR module leads to a performance drop of 0.88 BLEU and 0.82 METEOR. This highlights the necessity of retrieving globally distributed segments to capture long-range dependencies, which are crucial for maintaining narrative coherence in documentaries.

\textbf{Impact of Text-aware Selector.} Excluding the TVSS module reduces BLEU to 29.71. This suggests that while retrieval provides candidates, our adaptive selection and fusion strategy is essential for enriching visual-semantic density and preserving subtle yet meaningful cues.

\textbf{Joint Contribution.} Removing both modules results in the lowest performance (29.48 BLEU). This confirms that the global retrieval and adaptive selection mechanisms are complementary, jointly enabling the model to construct comprehensive and semantically aligned visual representations.

\subsection{Generalizability on Large Vision-Language Models}

To verify the scalability and generalizability of our method within the paradigm of large vision–language models (VLMs), we further evaluate the effectiveness of the selectively incorporating global video information strategy described above on Qwen3-VL~\cite{qwen3technicalreport}. 
Distinct from the standard uniform sampling strategy adopted by Qwen3-VL, we first explore simple alternatives, including dot-product similarity and a jointly learned selection module, to selectively filter video frames based on their textual relevance, thereby constructing a semantically concentrated visual context. 

We conduct validation experiments on Qwen3-VL-2B-Instruct~\footnote{\url{https://huggingface.co/Qwen/Qwen3-VL-2B-Instruct}}. The results indicate that the BLEU score improves from 26.70 for the base VLM to 27.75 when using dot-product similarity, and further increases to 28.63 with the jointly learned selection module.
These findings fully underscore the effectiveness of our method, demonstrating that the proposed text-aware frame selection mechanism substantially enhances the model's cross-modal alignment capabilities.

\section{Conclusions}

In this work, we propose a novel global context-aware multimodal translation framework that leverages semantic relationships across video segments to improve VMT. Unlike existing methods that rely mainly on locally aligned video--text pairs, our approach constructs a coherent, context-rich video segment collection by retrieving topically related subtitles via vector-based similarity search and expanding context with adjacent segments. 
The experimental results confirm the effectiveness of the proposed method, while the experiments on vision–language models further demonstrate the feasibility of our approach for leveraging global video information.
As future work, we plan to extend the global video context utilization strategies described in the preceding sections to VLMs, with the goal of enhancing their ability to process long video inputs.
%
%

\bibliographystyle{IEEEtran}
\bibliography{refs}

\end{document}